\documentclass[10pt, twocolumn, letterpaper]{article}
\usepackage{cvpr}              

\usepackage{colortbl}
\usepackage{graphicx}
\usepackage{amsmath}
\usepackage{amssymb}
\usepackage{booktabs}
\usepackage{graphicx}
\usepackage{multirow}
\usepackage{amsfonts,mathtools}
\usepackage{stfloats}
\usepackage{cuted}
\usepackage{flushend}
\usepackage{tcolorbox}

\definecolor{cvprblue}{rgb}{0.21,0.49,0.74}
\usepackage[pagebackref,breaklinks,colorlinks,allcolors=cvprblue]{hyperref}

\newcommand{\specialcell}[2][l]{%
  \begin{tabular}[#1]{@{}l@{}}#2\end{tabular}}

\title{DSV-LFS: Unifying LLM-Driven Semantic Cues with Visual Features for Robust Few-Shot Segmentation}

\author{Amin Karimi, Charalambos Poullis\\
Immersive and Creative Technologies Lab\\
Concordia University\\
Montreal, Quebec, Canada\\
{\tt\small aminpdi@gmail.com, charalambos@poullis.org}
}

\begin{document}
\maketitle
\begin{abstract}
 Few-shot semantic segmentation (FSS) aims to enable models to segment novel/unseen object classes using only a limited number of labeled examples. However, current FSS methods frequently struggle with generalization due to incomplete and biased feature representations, especially when support images do not capture the full appearance variability of the target class. To improve the FSS pipeline, we propose a novel framework that utilizes large language models (LLMs) to adapt general class semantic information to the query image. Furthermore, the framework employs dense pixel-wise matching to identify similarities between query and support images, resulting in enhanced FSS performance.
Inspired by reasoning-based segmentation frameworks, our method, named DSV-LFS, introduces an additional token into the LLM vocabulary, allowing a multimodal LLM to generate a "semantic prompt" from class descriptions. In parallel, a dense matching module identifies visual similarities between the query and support images, generating a "visual prompt". These prompts are then jointly employed to guide the prompt-based decoder for accurate segmentation of the query image. Comprehensive experiments on the benchmark datasets Pascal-$5^{i}$ and COCO-$20^{i}$ demonstrate that our framework achieves state-of-the-art performance-by a significant margin-demonstrating superior generalization to novel classes and robustness across diverse scenarios. The source code is available at \href{https://github.com/aminpdik/DSV-LFS}{https://github.com/aminpdik/DSV-LFS}
 
\end{abstract}

\section{Introduction}
\label{sec:intro}

Deep neural networks have shown remarkable success in learning visual features from large labeled datasets \cite{he2016deep,dosovitskiy2020image,YOLO,YOLOv2,FasterRCNN}. However, their ability to generalize to new classes diminishes when only a limited labeled data is available. Few-shot learning \cite{vinyals2016matching,finn2017model,snell2017prototypical} addresses this limitation by enabling models to learn effectively from a small number of labeled examples, similar to human learning.

In the context of image segmentation \cite{long2015fully,tian2023learning}, which requires pixel-level annotations, few-shot learning provides a resource-efficient solution. Few-shot semantic segmentation methods focus on predicting detailed masks for novel classes using only a limited number of labeled samples (support images). These methods utilize a range of strategies to effectively leverage the limited labeled samples available for segmentation \cite{shaban2017one,liu2020crnet,tian2020prior,zhang2019canet,wang2019panet,Okazawa2022Interclass,Hong2021VAT,REPPRI,Sun2022Singular,fan2022self,lang2022learning}. However, current methods face challenges such as overfitting to the feature distribution of the training data during meta-training, leading to misclassification of seen classes as unseen ones. Additionally, occlusion, deformation, or texture differences between query and support images significantly reduce segmentation accuracy. 
The root cause for these challenges is the incomplete and appearance-biased feature representation of the novel class learned from the limited data available. Recent methods address these challenges by leveraging class text descriptions, which provide detailed semantic information to improve segmentation performance \cite{zhu2023llafs,liu2023referring,yang2023mianet,ma2023attrseg}. These descriptions help models capture nuanced features of novel classes, enhancing generalization and accuracy even with limited support images. Advances in large language models (LLMs) \cite{touvron2023llama,brown2020language} further enable the efficient encoding of this semantic information, offering a more robust integration of textual and visual cues for improved segmentation.

Large language models (LLMs) have shown great potential in few-shot learning, enhancing performance across tasks in both computer vision and natural language processing \cite{brown2020language,touvron2023llama,brown2023multimodal,NEURIPS2022_8bb0d291,chen2023unified,liu2023crossmodal}. In few-shot segmentation (FSS), integrating LLMs to encode textual information has proven effective in addressing key limitations. While earlier FSS methods used language models for auxiliary tasks such as feature extraction \cite{xu2023self,liu2023referring,he2023primitive} or generating attribute prompts \cite{ma2023attrseg}, recent work \cite{zhu2023llafs} presents the first direct application of LLMs to FSS, achieving notable improvements in segmentation accuracy. \cite{zhu2023llafs} engineered prompts leveraging both the support set and class description to guide the LLM in performing segmentation on the query image.
However, this method has several key limitations, such as requiring multi-stage training and remaining text-centric, with segmentation results generated as textual descriptions that must be then post-processed to produce a segmentation mask.
Despite \cite{zhu2023llafs} having shown significant improvements through the direct application of LLMs in few-shot settings, an important challenge remains: developing a single-stage, end-to-end pipeline that leverages text-based LLMs to efficiently integrate support images and class descriptions for direct query image segmentation.

To generate segmentation directly on the query image, we adapted a prompt-based decoder \cite{kirillov2023segment} to harness its ability to integrate image features with user-provided prompts, facilitating the generation of accurate segmentation masks. These prompts guide the decoder in localizing the region of interest within the query image. Building on recent FSS methods, we efficiently utilize both the support set and class descriptions to guide the decoder.

However, generating prompts from class descriptions presents a significant challenge. While class descriptions provide consistent, general visual information about the object class, the current query image may lack some of these characteristics due to variations in appearance caused by occlusions, lighting conditions, or partial visibility. As a result, directly encoding class descriptions and incorporating them into the FSS pipeline is not efficient. To overcome this challenge, and inspired by the \textit{reasoning segmentation} framework \cite{lai2023lisa}, we introduce an additional token, $<SEM_{prompt}>$, into the LLM vocabulary, which signifies a request for segmentation based on semantic information. We further design a class semantic encoder module based on a multimodal LLM \cite{liu2023visual, liu2024improved}, which takes both the query image and general class description to generate query-specific semantic information, referred to as semantic prompt. To further enhance performance, our method incorporates a dense matching module that encodes the similarity between query and support images, producing a visual prompt. This visual prompt complements the semantic prompt by providing fine-grained spatial correspondence, enabling the decoder to better align the class-specific features with the query image. By combining these two forms of guidance, the adapted prompt-based decoder effectively mitigates the limitations of traditional FSS pipelines, delivering superior segmentation accuracy in challenging scenarios and achieving state-of-the-art results by a significant margin.

Our main contributions are as follows:
\begin{itemize}
    \item To the best of our knowledge, this is the first work, that combines large language models (LLMs) fine-tuned for reasoning segmentation \cite{lai2023lisa}, with foundation semantic segmentation models to directly segment in the context of few-shot semantic segmentation.
    \item We propose a novel single-stage, end-to-end architecture that seamlessly integrates multimodal semantic features from large language models with visual features derived from pixel-level correspondence, resulting in substantial improvements in segmentation accuracy and robustness.
    \item We conduct comprehensive experiments across multiple benchmark datasets, demonstrating that our method achieves state-of-the-art, outperforming existing methods by a significant margin.
\end{itemize}

\begin{figure*}[!ht]
    \centering
    \includegraphics[width=\textwidth]{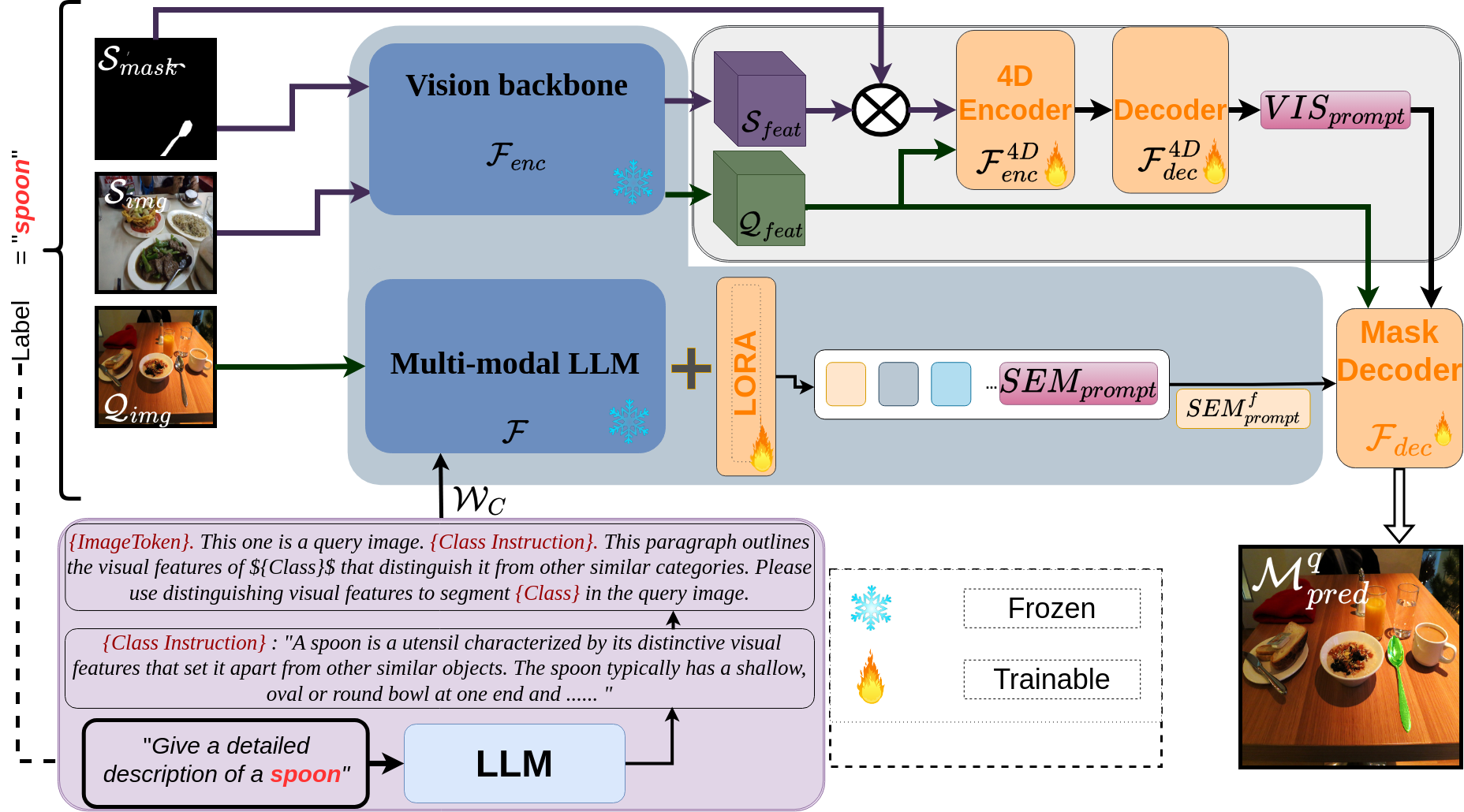} 
    \caption{\textbf{Technical Overview.} The large language model (LLM) first generates a class description $W_{C}$ based on an input prompt, which consists of a simple question regarding the visual features that distinctly define the class $C$ with label $\xi$. The $\{ImageToken\}$ in $W_{C}$ serves as a default token assigned to the query image, and $\{Class\}$ refers to the class label $\xi$. This class description, along with the query image, is then fed into a multi-modal LLM $(\mathcal{F})$ to produce a class-specific semantic prompt $SEM^{f}_{prompt}$. In parallel, a dense matching module $\mathcal{F}_{enc}^{4D}$, $\mathcal{F}_{dec}^{4D}$, generates a class-specific visual prompt $VIS_{prompt}$ by using the support and query feature maps obtained from the vision backbone encoder $\mathcal{F}_{enc}$. Finally, these two prompts, together with the query feature maps, are passed to the prompt-based decoder $\mathcal{F}_{dec}$ to produce the final segmentation.}
    \label{fig:teaser}
\end{figure*}

\section{Related Work}
\label{sec:related}

\textbf{Few-Shot Segmentation.} Classical semantic segmentation methods often rely on a large number of training samples to achieve accurate results. However, to reduce this dependency, few-shot segmentation has emerged as a promising alternative, enabling the segmentation of query images with only a few annotated support images. This approach has received significant attention in recent years. Most FSS methods rely on matching query and support images using prototype-based \cite{liu2020crnet,tian2020prior,Zhang2021self,wang2019panet,Okazawa2022Interclass,siam2019amp,nguyen2019feature,Yang2020PrototypeMixture} or pixel-wise methods \cite{min2021hypercorrelation,Hong2021VAT,Shi2022DCAMA,pixelmatch2022}. In prototype-based methods, prototypes are extracted from support images and used for segmentation using either parametric or non-parametric approaches. Non-parametric \cite{wang2019panet,dong2018few,li2021adaptive,Yang_2021_ICCV,zhang2020sgone} methods classify pixels in the query image based on their similarity to the support prototype, while parametric methods \cite{zhang2019canet,lang2022learning,liu2020crnet,wu2021learning,xie2021scaleaware} use learnable parameters to segment the query image using aggregated features of the support and query images. Prototype-based methods in FSS can result in a significant loss of information, as pixel-level features are averaged into a single prototype. To address this limitation, recent work has focused on learning pixel-wise correlations between query and support features, offering a more detailed and accurate approach to segmentation. However, these methods still primarily rely on the limited information available in support images, which may not be sufficient for robust segmentation.
To overcome these challenges, several techniques have been developed to incorporate additional information into the few-shot segmentation process. For instance, \cite{lang2022learning} utilizes a fully supervised, pre-trained network on seen classes to generate a prior map, reducing confusion between seen and unseen classes. \cite{yang2023mianet} enhances segmentation by incorporating class semantic information, encoding class names using word2vec models, and integrating this information into the query-support matching pipeline for better segmentation outcomes. \cite{zhu2023llafs}, directly employs LLMs for segmentation by introducing detailed task instructions and fine-grained in-context guidance, simulating human cognition to enhance LLMs ability to generate accurate segmentation by providing refined multimodal references.

\textbf{Large Language Models.}  LLMs have driven substantial advancements in machine learning, significantly altering the ways in which various tasks are addressed and solved \cite{touvron2023llama,touvron2023llama,brown2020language,zhao2023survey}. These models excel at generating text that closely emulates human language, and they demonstrate exceptional versatility across a wide range of tasks, including transfer learning, few-shot learning, and zero-shot learning \cite{brown2020language,NEURIPS2022_8bb0d291,zhao2023llmknowledge,brown2023multimodal,openai2023llmclassification,wei2022zeroshot}. Recently, a number of multi-modal LLM models have been proposed for tasks involving image reasoning, which require a deep integration of visual and textual information to enhance comprehension and interpretation of images. For instance, models like those in \cite{liu2023visual,liu2024improved} combine the strengths of large language models with advanced visual processing to manage tasks that require understanding both text and images, such as visual question answering and image captioning.  Similarly, LISA \cite{lai2023lisa} presents a model that incorporates large language models with image segmentation capabilities. This approach uses a $<SEG>$ token to encode input prompts, and the resulting last-layer embedding is decoded into a segmentation mask through the model decoder, leveraging visual features extracted by a vision backbone like SAM \cite{kirillov2023segment}. Our method improves the few-shot segmentation process by utilizing multimodal LLMs to adapt detailed object class descriptions dynamically to the query image. This semantic information is seamlessly incorporated into the query and support image matching pipeline to guide a prompt-based decoder, enabling more accurate segmentation.

\section{Method}
\label{sec:method}

\subsection{Problem Definition}
Few-shot segmentation focuses on segmenting a target object in a query image with the help of only a few annotated support images. This approach uses meta-learning, where the model is trained using episodes instead of conventional image batches. Each episode consists of a support set and a query set. In a K-shot setting, the support set $S= \{X_{i}^{s},M_{i}^{s}\}_{i=1}^{K}$ includes $K$ support images $X^{s}$ and their corresponding masks $M^{s}$, while the query set $Q=\{X^{q},M^{q}\}$ comprises a query image $X^{q}$ and its corresponding mask $M^{q}$ used for the loss calculation during training. 

In the standard few-shot segmentation setting, the support set provides annotated examples for a target class $C \in C_{train} \cup C_{test} $, inherently including the class label through the segmentation masks. Our method leverages this class label $C$ to generate a class description $W_{C}$ using ChatGPT, which enriches the semantic understanding of the target class without introducing additional information beyond what is available in traditional FSS tasks.
The episodes are drawn from the training dataset $D_{train}= \{(S_{i},Q_{i},W_{C})\}_{i=1}^{N_{train}}$ for the meta-training phase and from the testing dataset $D_{test}=\{(S_{i},Q_{i},W_{C})\}_{i=1}^{N_{test}}$ for meta-testing. $D_{train}$ has $C_{train}$ classes, and $D_{test}$ has $C_{test}$ classes, with no overlap between the two, i.e., $C_{train} \cap C_{test} = \emptyset$. The goal is to train the model on $D_{train}$ and then test it on unseen classes in $D_{test}$, leveraging the learn-to-learn paradigm.

\subsection{Overview}
The proposed few-shot semantic segmentation network comprises three core modules, as depicted in Figure \ref{fig:teaser}. First, the Class Semantic Encoder $\mathcal{F}$ employs a multimodal LLM to adapt the general class description to the query image, generating the $<SEM_{prompt}>$. Second, the Dense Matching Module identifies visual correspondences between the query and support images, producing the $VIS_{prompt}$. Finally, the prompt-based Decoder Module $\mathcal{F}_{dec}$ combines these two complementary prompts with query image features extracted by a visual encoder $\mathcal{F}_{enc}$ to accurately segment the target object.

\subsection{Class Description Generation}

To generate the class descriptions, we query ChatGPT 4.0 with a specific prompt for each class with label $\xi$ as follows: \textcolor{RoyalBlue}{Please provide a detailed description of the visual characteristics that uniquely identify the $<class name>$ object class, distinguishing it from other similar object categories. Focus solely on the distinguishing visual features in a comprehensive paragraph.} This prompt ensures that the descriptions are tailored to highlight the unique visual features of each class. We collect the responses from ChatGPT for all classes. 
As an example, the following is a class description generated for "spoon": \textcolor{RoyalBlue}{A spoon is a utensil characterized by its distinctive visual features that set it apart from other similar objects. The spoon typically has a shallow, oval or round bowl at one end, designed to hold and scoop liquids or semi-solids. The handle of a spoon is usually elongated, straight, or slightly curved, and tapers towards the bowl, allowing for a comfortable grip. They are commonly made from reflective materials like stainless steel, which give them a shiny appearance, but can also be found in other materials such as wood or plastic, each presenting a different texture and finish. Unlike forks, spoons lack tines, and unlike knives, they do not have a blade or sharp edges, making their overall form smooth and rounded.}

We experimented with various prompt formulations to mitigate sensitivity in LLM outputs. Our final prompt structure described above emphasizes distinctive visual features, and we found this approach to consistently generate informative and reliable class descriptions.

Incorporating the generated class descriptions, and following the approach of \cite{liu2023visual,liu2024improved}, we construct input prompt for class encoder module ($\mathcal{W}_{C}$ in Figure \ref{fig:teaser}) as: \textcolor{RoyalBlue}{${ImageToken}$. This one is a query image. ${Class Instruction}$. This paragraph outlines the visual features of ${Class}$ that distinguish it from other similar categories. Please use distinguishing visual features to segment ${Class}$ in the query image.}
The ${ImageToken}$ serves as a default token assigned to the query image within the input prompt. This token is subsequently replaced by the output features of the query image generated by CLIP \cite{radford2021learning}. The ${Class Instruction}$ corresponds to the specific description generated for the target class, while the ${Class}$ represents the name of the object category being segmented. The expected output follows this format: \textcolor{RoyalBlue}{Sure, the segmentation result is $<SEM_{prompt}>$}, where $<SEM_{prompt}>$ is a token added to the LLM vocabulary. This token enables the LLM to adapt the general class description to the visual features of the target object, providing query image-specific semantic information.

\subsection{Class Semantic Encoder Module}

Class descriptions offer general visual information about an object class, but query images often vary due to factors such as occlusions, lighting changes, or partial visibility. As a result, directly incorporating class descriptions into the FSS pipeline is not efficient. To address this, we propose a class semantic encoder module that adapts class descriptions to the query image, generating a context-aware semantic prompt that captures the specific characteristics of the target class within the query image. To enable this adaptation, we introduce the $<SEM_{prompt}>$ token into the LLM vocabulary, which requests detailed, query image-specific semantic information.
Following the LLaVA architecture, features from the query image $Q_{img}^{\mathcal{C}}$ containing the target object class $\mathcal{C}$, extracted by a visual encoder, i.e., CLIP \cite{radford2021learning}, along with the prepared description $W_{\mathcal{C}}$ for the target object class, are fed into the LLM, which generates a text response.

\begin{align}
    & y_{text}= \mathcal{F} (Q_{img}^{C},W_{\mathcal{C}})
\end{align}

When instructed to generate a segmentation aligned with the target class description, the output $y_{text}$ includes the $<SEM_{prompt}>$ token, which encodes semantic information specific to the target class in the context of the query image.
 Next, we extract the LLM last-layer embedding, $h_{sem}$ , corresponding to the $<SEM_{prompt}>$ token, and apply an MLP projection layer to obtain the final semantic prompt  $SEM^{f}_{prompt}$.
\begin{align}
    & SEM^{f}_{prompt}= MLP_{proj}(h_{sem})
\end{align}

\subsection{Dense Matching Module}
This module generates a visual prompt $VIS_{prompt}$ that encodes the similarity between the target object in the support set and the query image. To achieve this, we leverage dense matching between annotated support images and the query image, which has been shown to outperform prototype-based matching by capturing fine-grained details \cite{min2021hypercorrelation}.
We extract a diverse range of features from various $L$ depths  of a vision transformer, i.e., the SAM  \cite{kirillov2023segment}, $\{(f_{S}^{l}, f_{Q}^{l})\}_{l=1}^{L}$ forming a set of 4D hypercorrelation tensors, $HPV_{l} \in \mathbb{R}^{H_{p} \times W_{p} \times H_{p} \times W_{p}}$. These 4D hypercorrelations are then stacked together.
\begin{align}
    & HPV_{l_{(X_q, X_s)}}= \text{ReLU} \left( \frac{f_{Q}^{l} \cdot f_{S}^{l}}{\|f_{Q}^{l}\| \|f_{S}^{l}\|} \right)
\end{align}
\begin{equation}  
\begin{aligned}
    \textbf{HPV} &= \text{Concat} \Big( HPV_{1}, HPV_{2}, \ldots, HPV_{L} \Big)
\end{aligned}
\end{equation}
\begin{equation}  
\begin{aligned}
    \textbf{HPV} &\in \mathbb{R}^{L \times H_{p} \times W_{p} \times H_{p} \times W_{p}}
\end{aligned}
\end{equation}

Using efficient center-pivot 4D convolutions 
($\mathcal{F}^{4D}_{enc}$ in Figure \ref{fig:teaser}), the method combines high-level semantic and low-level geometric cues from hypercorrelations to encode matching of support and query images. The encoder output is then passed to a decoder module ($\mathcal{F}^{4D}_{dec}$ in Figure \ref{fig:teaser}), which generates a visual prompt, $VIS_{prompt}$, to guide the segmentation process.
\begin{align}
    & \mathcal{H}^{4D}= \mathcal{F}^{4D}_{enc} (\textbf{HPV})
\end{align}
\begin{align}
    & VIS_{prompt}= \mathcal{F}^{4D}_{dec} (\mathcal{H}^{4D})
\end{align}


\subsection{Mask Decoder Module}

In our method, we employ two types of prompts—semantic ($SEM^{f}_{prompt}$) and visual ($VIS_{prompt}$)—to guide the prompt-based decoder ($\mathcal{F}_{dec}$ in Figure \ref{fig:teaser}) for accurate segmentation of the query image. The semantic prompt, produced by the Class Semantic Encoder, captures detailed attributes of the target object, such as shape, texture, and distinctive features, providing contextual information for segmentation. The visual prompt, generated by the Dense Matching Module, is derived from pixel-wise matching between the query image and annotated support images, identifying the target regions in the query image. These prompts, combined with query image features $Q_{feat}$, extracted by the encoder $\mathcal{F}_{enc}$, are integrated into the prompt-based decoder $\mathcal{F}_{dec}$ \cite{kirillov2023segment} to directly generate segmentation on the query image.
\begin{align}
    & M^q_{pred}=\mathcal{F}_{dec}(Q_{feat},VIS_{prompt},SEM^{f}_{prompt})
\end{align}

The decoder employs multi-head attention and transformer blocks to fuse the semantic and visual information, refining mask proposals through learnable queries. This hierarchical approach merges the high-level context from the semantic prompt with spatial details from the visual prompt, producing detailed segmentation masks. This integration enables the decoder $\mathcal{F}_{dec}$ to deliver segmentation results that are precise and contextually coherent.

\subsection{Training loss}
Our model is trained end-to-end using two main loss functions adapted from \cite{lai2023lisa}: the text generation loss \(\mathcal{L}_{\text{text}}\) and the segmentation mask loss \(\mathcal{L}_{\text{mask}}\). The overall loss function \(\mathcal{L}\) is defined as the weighted sum of these two losses:
\begin{align}
\mathcal{L} = \lambda_{\text{text}} \mathcal{L}_{\text{text}} + \lambda_{\text{mask}} \mathcal{L}_{\text{mask}},
\end{align}

where \(\lambda_{\text{text}}\) and \(\lambda_{\text{mask}}\) are the weights assigned to the text and mask losses, respectively. The text generation loss \(\mathcal{L}_{\text{text}}\) is formulated as the auto-regressive cross-entropy loss, while the segmentation mask loss \(\mathcal{L}_{\text{mask}}\) combines per-pixel binary cross-entropy (BCE) loss and Dice loss, weighted by \(\lambda_{\text{BCE}}\) and \(\lambda_{\text{Dice}}\).

Given the ground truth text labels \(\mathbf{\hat{y}}_{\text{text}}\) and query mask \(\mathbf{M}^{q}\) and predicted mask for query image \(\mathbf{M}^{q}_{pred}\), the specific loss functions are:
\begin{align}
\mathcal{L}_{\text{text}} &= \text{CE}(\hat{\mathbf{y}}_{\text{text}}, \mathbf{y}_{\text{text}})
\end{align}
\begin{align}
\mathcal{L}_{\text{mask}} = &\lambda_{\text{BCE}} \text{BCE}(\mathbf{M}^{q}_{pred}, \mathbf{M}^{q}) + \\
&\lambda_{\text{Dice}} \text{Dice}(\mathbf{M}^{q}_{pred}, \mathbf{M}^{q})
\end{align}
Our method integrates these losses to extend the capabilities of multimodal large language models (LLMs), enabling them to handle both text generation and fine-grained segmentation tasks. 

\subsection{Extending to $K$-shot setting}
For the $K$-shot scenario, we adopt the strategy proposed in \cite{min2021hypercorrelation}. With $K$ support image-mask pairs and a query image, the model makes $K$ separate forward passes, resulting in $K$ predicted segmentation masks. To determine the final segmentation, a voting mechanism is applied at each pixel, where the sum of the $K$ predictions is normalized by the maximum possible votes. Pixels are then classified as foreground if their normalized score exceeds a certain threshold, allowing for a more robust segmentation decision based on multiple support examples.

\begin{table*}[ht!]
\centering
\caption{\textbf{Performance Comparisons.} We evaluate our method by comparing the mean intersection-over-union (mIoU) on the PASCAL-$5^{i}$ and COCO-$20^{i}$ datasets against other state-of-the-art methods. To ensure the robustness and reliability of the results, we perform each experiment five times using different random seeds and report the average mIoU scores for both 1-shot and 5-shot settings. The highest values are indicated in \textbf{bold}, the second-highest are \underline{underlined}, and the average mIoU is \colorbox{gray!70}{highlighted}.
}
 \label{tab:Compresults}
\resizebox{\textwidth}{!}{
\begin{tabular}{@{}p{1.5cm}|p{2.5cm}|p{1.7cm}|c |c |c |c ||c| c |c |c |c ||c@{}}
\toprule
\specialcell[t]{\textbf{Dataset}} & \specialcell[t]{\textbf{Method}} & \specialcell[t]{\textbf{Conference}}& \multicolumn{5}{c|}{\cellcolor[gray]{0.9}\textbf{1-shot}} & \multicolumn{5}{c}{\cellcolor[gray]{0.9}\textbf{5-shot}} \\ 
&  &  & Fold-0 & Fold-1 & Fold-2 & Fold-3 & Mean & Fold-0 & Fold-1 & Fold-2 & Fold-3 & Mean \\
\midrule
\multirow{13}{*}{PASCAL-$5^{i}$} & NTRENet \cite{liu2022learning} & CVPR'22 & 65.4 & 72.3 & 59.4 & 59.8 & 63.2 & 66.2 & 72.8 & 61.7 & 62.2 & 65.7 \\

& BAM \cite{lang2022learning} & CVPR'22 & 69.0 & 73.6 & 67.5 & 61.1 & 67.8 & 70.6 & 75.1 & 70.8 & 67.2 & 70.9 \\

& AAFormer \cite{wang2022adaptive} & ECCV'22 & 69.1 & 73.3 & 59.2 & 65.2 & 66.7 & 72.5 & 74.7 & 62.0 & 61.3 & 67.6 \\

& SSP \cite{fan2022self} & ECCV'22 & 60.5 & 67.8 & 56.1 & 61.4 & 61.5 & 67.5 & 72.7 & 75.2 & 62.1 & 69.3 \\

& IPMT \cite{liu2022intermediate} & NeurIPS'22 & 72.8 & 73.7 & 59.2 & 61.6 & 66.8 & 73.1 & 74.7 & 61.6 & 63.4 & 68.2 \\

& ABCNet \cite{wang2023rethinking} & CVPR'23 & 68.8 & 73.4 & 59.6 & 65.0 & 66.5 & 71.7 & 74.2 & 74.8 & 67.0 & 69.6 \\

& HDMNet \cite{peng2023hierarchical} & CVPR'23 & 71.0 & 75.4 & 62.1 & 69.4 & 69.7 & 71.3 & 76.2 & 71.3 & 68.5 & 71.8 \\

& MIANet \cite{yang2023mianet} & CVPR'23 & 68.5 & 75.8 & 64.5 & 68.7 & 69.4 & 70.2 & 77.4 & 70.0 & 68.8 & 71.7 \\

& MSI \cite{moon2023msi} & ICCV'23 & 71.0 & 72.5 & 63.8 & 65.9 & 68.3 & 73.0 & 74.2 & 70.5 & 66.6 & 71.1 \\

& SCCAN \cite{xu2023self} & ICCV'23 & 68.3 & 72.5 & 66.8 & 58.9 & 66.6 & 72.3 & 74.1 & 69.1 & 65.6 & 70.3 \\

& LLaFS \cite{zhu2023llafs}& CVPR’24 & \textbf{74.2} & \underline{78.8} & \textbf{72.3} & \underline{68.5} & \underline{73.5} & \textbf{75.9} & \underline{80.1} & \textbf{75.8} & \underline{70.7} & \textbf{75.6} \\

\rowcolor[HTML]{D9EAD3} 
& \textbf{DSV-LFS} &  & \underline{71.67} & \textbf{81.97} & \underline{71.17} &  \textbf{75.04} &\cellcolor[gray]{0.7} \textbf{74.96} & \underline{72.03} & \textbf{82.01} & \underline{71.32} & \textbf{75.51} & \cellcolor[gray]{0.7} \underline{75.21} \\
\hline
\hline
\multirow{13}{*}{COCO-$20^{i}$} & NTRENet \cite{liu2022learning} & CVPR'22 & 36.8 & 42.6 & 39.7 & 39.3 & 38.2 & 38.2 & 44.1 & 40.4 & 38.4 & 40.3 \\

& BAM \cite{lang2022learning} & CVPR'22 & 43.4 & 50.6 & 47.5 & 43.6 & 46.3 & 49.3 & 54.2 & 51.6 & 49.9 & 51.2 \\

& SSP \cite{fan2022self} & ECCV'22 & 35.5 & 39.6 & 37.9 & 36.7 & 37.4 & 40.6 & 47.0 & 45.1 & 43.9 & 44.1 \\

& AAFormer \cite{wang2022adaptive} & ECCV'22 & 39.8 & 44.6 & 41.1 & 41.6 & 41.8 & 42.9 & 50.1 & 45.5 & 49.6 & 49.6 \\

& MM-Former \cite{zhang2022mask} & NeurIPS'22 & 40.5 & 47.7 & 45.2 & 43.4 & 44.2 & 40.4 & 47.4 & 50.0 & 48.8 & 46.6 \\

& IPMT \cite{liu2022intermediate} & NeurIPS'22 & 41.4 & 45.2 & 45.6 & 40.4 & 43.2 & 43.3 & 47.5 & 43.8 & 42.5 & 44.3 \\

& ABCNet \cite{wang2023rethinking} & CVPR'23 & 42.3 & 46.2 & 46.0 & 44.1 & 44.7 & 44.5 & 51.7 & 52.2 & 46.4 & 49.1 \\

& HDMNet \cite{peng2023hierarchical} & CVPR'23 & 43.8 & 50.8 & 50.6 & 49.4 & 48.6 & 50.6 & 61.6 & 55.7 & 56.6 & 56.1 \\

& MIANet \cite{yang2023mianet} & CVPR'23 & 42.5 & 50.3 & 47.8 & 47.4 & 47.7 & 45.8 & 58.2 & 51.3 & 51.9 & 51.7 \\

& MSI \cite{moon2023msi} & ICCV'23 & 42.4 & 47.4 & 44.9 & 44.6 & 44.8 & 47.1 & 53.2 & 53.4 & 51.9 & 51.9 \\

& SCCAN \cite{xu2023self} & ICCV'23 & 40.4 & 42.6 & 41.4 & 40.7 & 41.3 & 47.2 & 57.2 & 59.2 & 52.1 & 53.9 \\

& LLaFS \cite{zhu2023llafs}& CVPR’24 & \underline{47.5} & \underline{58.8} & \underline{56.2} & \underline{53.0} & \underline{53.9} & \underline{53.2} & \underline{63.8} & \underline{63.1} & \underline{60.0} & \underline{60.0} \\
\rowcolor[HTML]{D9EAD3} 
& \textbf{DSV-LFS} &  & \textbf{69.97} & \textbf{73.35} & \textbf{70.69} & \textbf{71.32} &\cellcolor[gray]{0.7} \textbf{71.33} & \textbf{71.05} & \textbf{73.81} & \textbf{71.32} & \textbf{71.45} & \cellcolor[gray]{0.7} \textbf{71.90} \\
\bottomrule
\end{tabular}}%
\label{tab:performance_comparison}
\end{table*}

\section{Experiments}
\label{sec:experiments}

\subsection{Experimental Settings}
\textbf{Benchmark datasets. } Following previous works in FSS, we evaluate the proposed method on two  few-shot segmentation benchmark datasets: Pascal-$5^{i}$ and COCO-$20^{i}$, derived from PASCAL VOC 2012 and MS-COCO, respectively. Each dataset is divided into four folds, with three-quarters of the classes designated for training (base/seen classes) and the remainder for testing (novel/unseen classes). During the inference phase, 1000 episodes of support and query images are randomly sampled from the test set to evaluate the model performance.

\noindent 
\textbf{Evaluation measures.} To evaluate the proposed method, we adopt mean intersection-over-union (mIoU), consistent with previous studies. To ensure robust and reliable results, we conduct five trials for each experiment using different random seeds. The final performance metric is obtained by averaging the results from all five trials, providing a comprehensive assessment of the method effectiveness for few-shot segmentation tasks.

\noindent 
\textbf{Implementation details.} The proposed network combines the pre-trained multi-modal language model LLaVA (llava-v1.5-7b) \cite{liu2024improved} with the Segment Anything network \cite{kirillov2023segment}. The network introduces a 4D dense matching module, which utilizes center-pivot 4D convolutions \cite{min2021hypercorrelation} followed by a convolutional decoder module. The mask decoder is derived from the Segment Anything mask decoder \cite{kirillov2023segment}. To generate class descriptions, a custom Python web scraping tool is used to query ChatGPT 4.0, producing detailed descriptions for all classes in the benchmark datasets.

\noindent 
\textbf{Training.} One advantage of the proposed model is that it functions as an end-to-end model trained in a single stage. To efficiently fine-tune the multi-modal LLM, we employ LoRA \cite{hu2021lora} while keeping the vision backbones frozen. Meanwhile, the 4D encoder/decoder and mask decoder are fine-tuned, and the LLM token embeddings, language model head (LM head), and projection layer are set as trainable parameters. The batch size is set to 2 per GPU,  and the model is trained for 10 epochs using the AdamW optimizer with the cosine annealing scheduler and an initial learning rate of $3e-4$. The loss weights are set to 1, 2, and 0.5 for $ \lambda_{\text{text}}, \lambda_{\text{BCE}}, \lambda_{\text{Dice}}$, respectively. To ensure a fair comparison with other methods, no data augmentation is used during training. Two NVIDIA A100 GPUs are employed for training.

\noindent 
\textbf{Fairness in Comparisons.} While our method incorporates detailed class descriptions generated from the class label, we maintain consistency with the standard FSS setting where the class label is inherently available through the support set annotations. Previous works have also leveraged class semantics, such as class names or word embeddings \cite{yang2023mianet, xian2019semantic} and more recently language guidance \cite{zhu2023llafs, wang2024language}, to enhance segmentation performance in FSS. 

\subsection{Comparison with State-of-the-Art}

\textbf{Pascal-$5^{i}$}: Table \ref{tab:Compresults} presents a comparison of the mIoU measure between our method and several recent few-shot segmentation approaches on the Pascal-$5^{i}$ benchmark. The results demonstrate that while our method significantly surpasses all non-LLM-based approaches, it achieves comparable results with the LLM-based approach of \cite{zhu2023llafs}. We attribute this to the dataset being simpler than other benchmarks, as it contains fewer classes and only one object class per image, resulting in potential saturation. This hypothesis is further supported by the results on other benchmark datasets, which surpass \textit{all} other methods by a significant margin, as we describe below.


\textbf{COCO-$20^{i}$:} Table \ref{tab:Compresults} presents the performance of our proposed method on the COCO-$20^{i}$ dataset, which is known for its challenging segmentation tasks due to the presence of multiple objects and significant intra-class variability. Our method achieves notable improvements over existing approaches, with gains of +17.43\% mIoU in the 1-shot setting and +11.9\% mIoU in the 5-shot setting.

\textbf{COCO-$20^{i}$ $\rightarrow$ Pascal-$5^{i}$}: On the basis of the different distributions of the training dataset and testing dataset, a model trained on one dataset is evaluated on another without any fine-tuning. To demonstrate the effectiveness of our method, we perform experiments in the COCO-$20^{i}$ $\rightarrow$ Pascal-$5^{i}$. We trained our network on all classes of COCO-$20^{i}$ and evaluate the network on Pascal-$5^{i}$ without fine-tuning. As shown in Table \ref{tab:coco_pascal}, while our network is not specifically designed for cross-domain few-shot segmentation, it still achieves state-of-the-art results, demonstrating a gain of +1\% mIoU in the 1-shot setting compared to other cross-domain FSS methods that are explicitly developed for this purpose.

\begin{figure*}[ht]
    \centering
    \renewcommand{\arraystretch}{0.5} 
    \resizebox{\textwidth}{!}{ 
    \begin{tabular}{@{} c @{} c @{} c @{} c @{} c @{} c @{} c @{} c @{}}
       &\parbox[c]{0.6cm}{\centering {\scriptsize \makebox[.6cm][c]{Motorcycle}}}&\parbox[c]{0.6cm}{\centering {\scriptsize \makebox[.6cm][c]{Train}}} &\parbox[c]{0.6cm}{\centering {\scriptsize \makebox[.6cm][c]{Potted plant}}} &\parbox[c]{0.6cm}{\centering {\scriptsize \makebox[.6cm][c]{handbag}}} &\parbox[c]{0.6cm}{\centering {\scriptsize \makebox[.6cm][c]{Laptop}}}&\parbox[c]{0.6cm}{\centering {\scriptsize \makebox[.6cm][c]{Fire hydrant}}}&\parbox[c]{0.6cm}{\centering {\scriptsize \makebox[.6cm][c]{Bird}}} \\ 
        
        \rotatebox{90}{\parbox[c]{0.6cm}{\centering \textcolor{ForestGreen}{\scriptsize \makebox[1.8cm][c]{Support Image}}}} &
        \includegraphics[width=2.0cm]{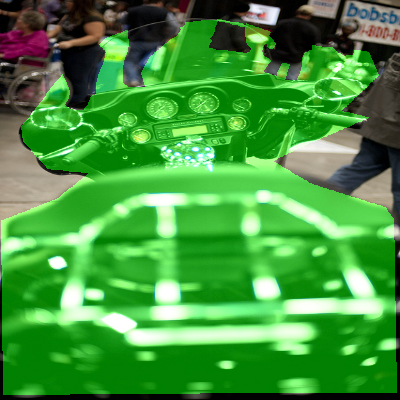} &  
        \includegraphics[width=2.0cm]{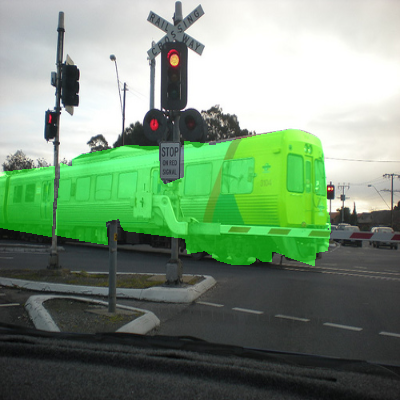} &
        \includegraphics[width=2.0cm]{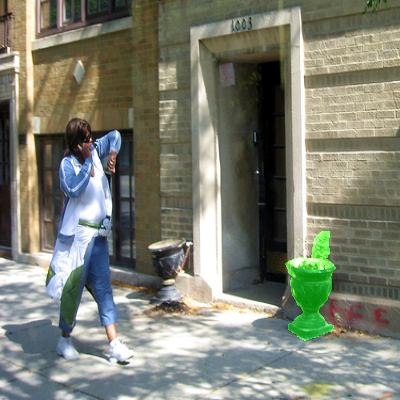} &
        \includegraphics[width=2.0cm]{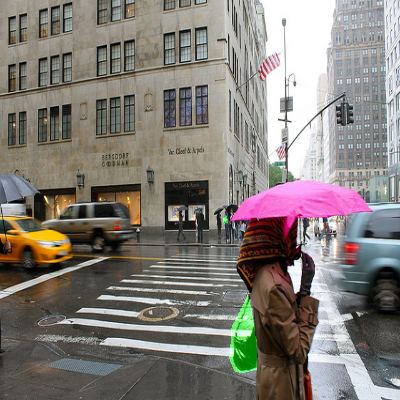} &
        \includegraphics[width=2.0cm]{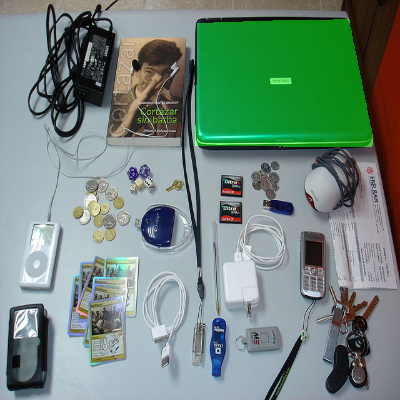} &
        \includegraphics[width=2.0cm]{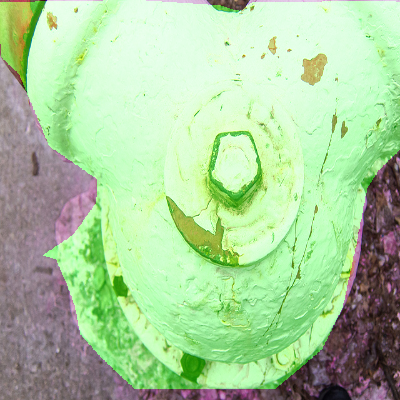} &
        \includegraphics[width=2.0cm]{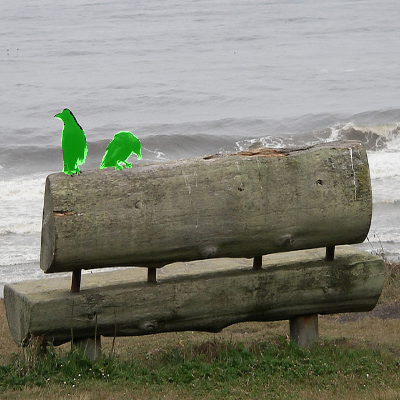} \\

        \rotatebox{90}{\parbox[c]{0.6cm}{\centering \textcolor{red}{\scriptsize \makebox[1.5cm][c]{Query Image}}}} &
        \includegraphics[width=2.0cm]{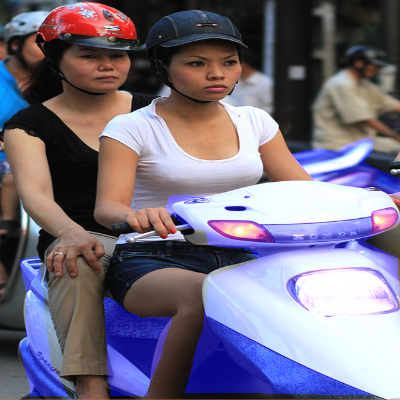} &  
        \includegraphics[width=2.0cm]{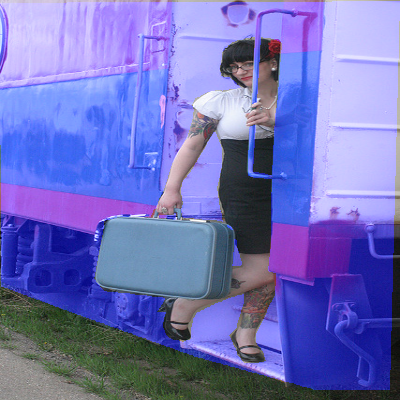} &
        \includegraphics[width=2.0cm]{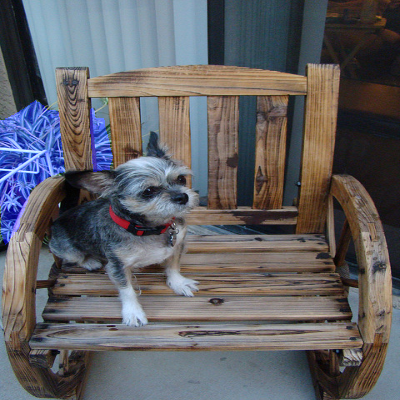} &
        \includegraphics[width=2.0cm]{figures/WACVFIG/8/Qovrly.png} &
        \includegraphics[width=2.0cm]{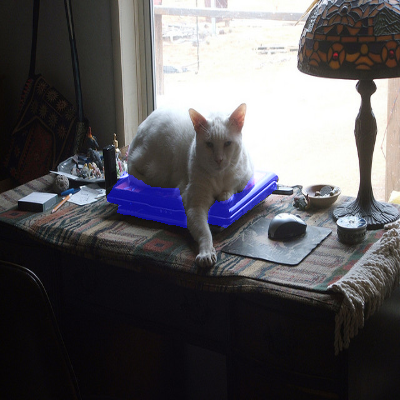} &
        \includegraphics[width=2.0cm]{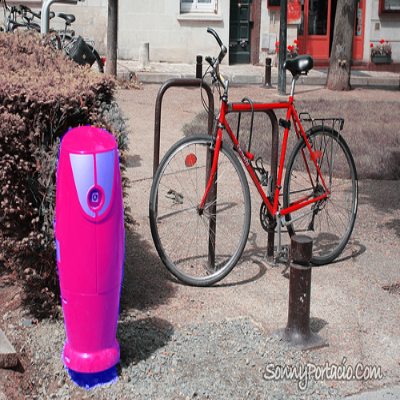} &
        \includegraphics[width=2.0cm]{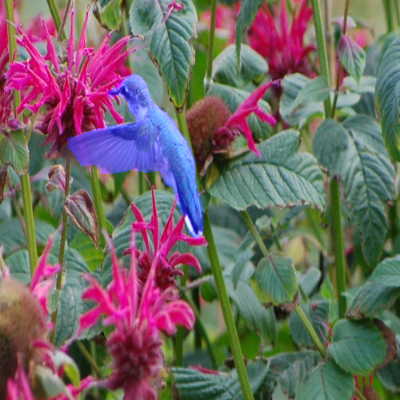} \\

        \rotatebox{90}{\parbox[c]{0.6cm}{\centering \scriptsize \makebox[1.5cm][c]{DSV-LFS}}} &
        \includegraphics[width=2.0cm]{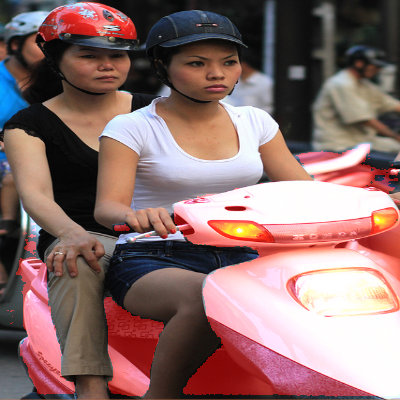} &  
        \includegraphics[width=2.0cm]{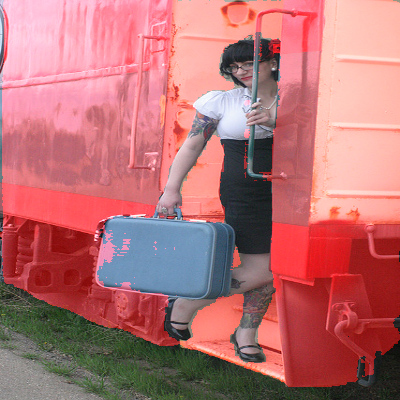} &
        \includegraphics[width=2.0cm]{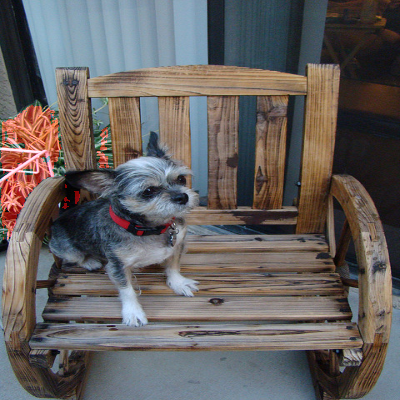} &
        \includegraphics[width=2.0cm]{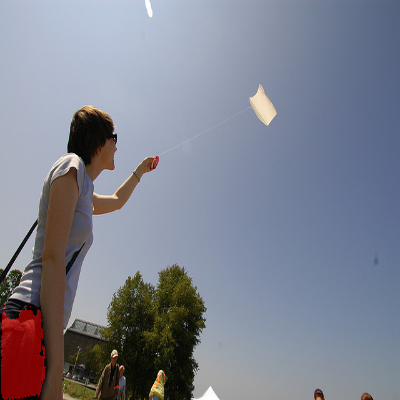} &
        \includegraphics[width=2.0cm]{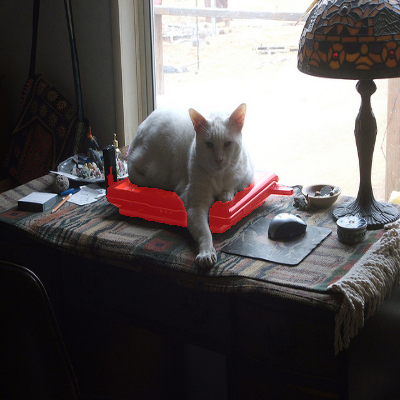} &
        \includegraphics[width=2.0cm]{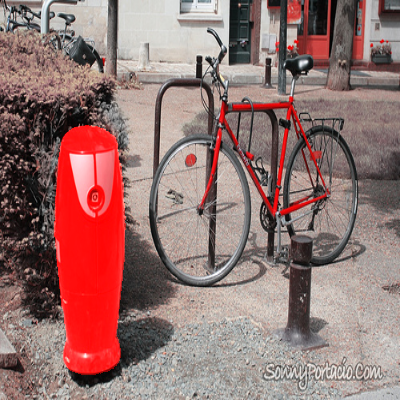} &
        \includegraphics[width=2.0cm]{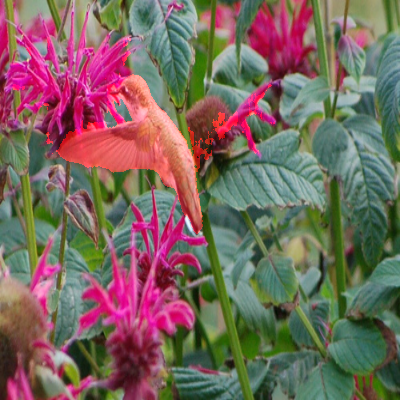} \\

    \end{tabular}
    } 
    \caption{\textbf{Qualitative results.}  Examples of our method's performance on the COCO-$20^{i}$ dataset. Each column represents an episode, displaying the support image, query image, and predicted segmentation output from top to bottom. The episodes illustrate the model's ability to handle challenges such as the presence of base classes in the query image (e.g., person in motorcycle and train classes) and variations between target objects in support and query images, including scale differences (e.g., handbag), occlusion (e.g., laptop), appearance changes (e.g., potted plant), complex backgrounds (e.g., bird), and deformations (e.g., fire hydrant).}
    \label{fig:results}
\end{figure*}
\subsection{Qualitative Results}

We present qualitative results generated by our proposed method to illustrate its effectiveness in overcoming key challenges in few-shot segmentation. Two common issues in this task include: (1) the misclassification of base class objects as novel classes, leading to false positives; and (2) the reliance on a few support images, which often do not capture the full variability of the target class appearance. Figure \ref{fig:results} demonstrates our method's robustness against misclassification of base classes as novel classes. The first and second columns highlight how our method accurately distinguishes target classes such as "motorcycle" and "train" from other objects in the query images. In the subsequent columns, we showcase the method's ability to precisely segment target classes despite significant variations, including differences in scale (e.g., handbag), occlusion (e.g., laptop), appearance changes (e.g., potted plant), cluttered backgrounds (e.g., bird), and deformations (e.g., fire hydrant) between the support and query images. These results underline the adaptability of our method to handle complex and diverse visual scenarios, substantially improving segmentation performance in few-shot learning contexts.
\begin{table}[ht]
\centering
\caption{
\textbf{Performance comparison of our model on the \textbf{COCO-$20^{i}$ $\rightarrow$ Pascal-$5^{i}$} cross-domain setting, without fine-tuning.} Although our method was not explicitly designed for cross-domain few-shot segmentation, it achieves state-of-the-art results with a +1 mIoU gain in the 1-shot setting. We run each experiment five times with different random seeds and report the average mIoU for the 1-shot setting. The highest values are indicated in \textbf{bold}, the second-highest are \underline{underlined}, and the average mIoU is \colorbox{gray!70}{highlighted}.
}
\label{tab:coco_pascal}
\resizebox{\linewidth}{!}{ 
\begin{tabular}{l|c|c|c|c|c}
\toprule
\multicolumn{6}{c}{\cellcolor[gray]{0.9}\textbf{COCO-20\textsuperscript{i} $\rightarrow$ Pascal-5\textsuperscript{i}}} 
\\
\textbf{Methods} & \textbf{Fold-0} & \textbf{Fold-1} & \textbf{Fold-2} & \textbf{Fold-3} & \textbf{mIoU} \\ 
\midrule
PFENet(TPMAI) \cite{tian2020prior} & 43.2 & 65.1 & 66.5 & 69.7 & 61.1 \\ 
RePRI(CVPR'21) \cite{REPPRI} & 52.2 & 64.3 & 64.8 & 71.6 & 63.2 \\ 
VAT(ECCV'22) \cite{Hong2021VAT} & 52.1 & 64.1 & 67.4 & 74.2 & 64.5 \\ 
VAT-HM(ECCV'22) \cite{Liu2022optimal} & 48.3 & 68.6 & 69.6 & \underline{79.8} & 65.6 \\ 
HSNet(ICCV'21) \cite{min2021hypercorrelation} & 47.0 & 65.2 & 67.1 & 77.1 & 64.1 \\ 
HSNet-HM(ECCV'22)\cite{Liu2022optimal} & 46.7 & 68.6 & \underline{71.1} & 79.7 & 66.5 \\ 
RTD(CVPR'22) \cite{Wang2022Meta} & 59.4 & 70.4 & 70.5 & 78.4 & 69.7 \\ 
PMNet (WACV'24) \cite{pixelmatch2022}  & \underline{71.0} & \underline{72.3} & 66.6 & 63.8 & 68.4 \\ 
IFA(CVPR'24) \cite{Nie_2024_CVPR} & - & - & - & - & \underline{79.6}\\ 
\rowcolor[HTML]{D9EAD3} 
\textbf{DSV-LFS} & \textbf{74.86} & \textbf{85.23} & \textbf{82.23} & \textbf{80.37} & \cellcolor[gray]{0.7}\textbf{80.67} \\ 
\bottomrule
\end{tabular}
}

\end{table}
\subsection{Ablations}
To assess the effectiveness of various components and design choices in our method, we conducted extensive ablation studies using the 1-shot setting of the COCO-$20^{i}$ dataset. We chose COCO-$20^{i}$ for our ablation study because it is a more challenging dataset with multiple objects per image and significant variability in object scales, poses, and contexts. This complexity makes it ideal for evaluating the robustness of each component of our proposed method. Table \ref{tab:ablation} illustrates how each component contributes to the overall model performance.

\begin{table}[ht]
\centering
\caption{\textbf{Ablation.} We evaluate segmentation performance using semantic prompts alone versus a combination of visual and semantic prompts by measuring the mean intersection-over-union (mIoU) on the \textbf{COCO}-$20^{i}$ dataset. The highest values are indicated in \textbf{bold}.}
\label{tab:ablation}
\resizebox{\linewidth}{!}{ 
\begin{tabular}{p{2cm}|p{1cm} | p{1cm} | p{1cm} | p{1cm} | p{1cm}} 
\toprule
\multirow{2}{*}{\textbf{Method}} & \multicolumn{5}{c}{\cellcolor[gray]{0.9}\textbf{1-shot}} \\ 
\cline{2-6}
  &   Fold-0 & Fold-1 & Fold-2 & Fold-3 & Mean  \\
\midrule
DSV-LFS w/ semantic prompt only &  66.99 & 71.34 & 67.52 & 70.14 & 68.99 \\
\hline
DSV-LFS w/ semantic \& visual prompt &  \textbf{69.97} & \textbf{73.35} & \textbf{70.69} & \textbf{71.32} & \textbf{71.33} \\
\bottomrule
\end{tabular}
}
\end{table}

\subsubsection{Effect of semantic prompt}

To assess the effectiveness of class descriptions, we conducted an experiment in which the mask decoder was guided exclusively by the semantic prompt generated by the Class Semantic Encoder Module. As shown in Table \ref{tab:ablation}, our method achieved a significant improvement of +15 mIoU over the current state-of-the-art FSS methods, even when relying solely on the semantic prompt. These results underscore the value of leveraging semantic knowledge from large language models in few-shot segmentation tasks.

\subsubsection{Effect of visual prompt}

An additional ablation was conducted to evaluate the effect of incorporating a small number of labeled samples alongside semantic information. In this experiment, we combined both the visual prompt from support images and the semantic prompt generated from class descriptions. The integration of these two prompts resulted in an approximate 3\% improvement as shown in Table \ref{tab:ablation}  compared to using the semantic prompt alone. This demonstrates that the visual prompt effectively complements the FSS pipeline by providing important visual cues that enhance the model ability to generalize to novel classes. The synergy between visual and semantic prompts indicates that leveraging both modalities can more effectively capture the diverse features and variations of target objects, thereby improving overall segmentation accuracy.

\section{Conclusion}
 This paper presents a novel approach to few-shot semantic segmentation that uniquely integrates LLM-derived semantic prompts with dense visual matching.
We introduce a new token, $<SEM_{prompt}>$, into the LLM vocabulary to generate class-specific semantic prompts, which are combined with visual prompts $<VIS_{prompt}>$ obtained through dense visual matching between query and support images. This dual-prompt strategy is inspired by the way the human brain and visual system rapidly learn and recognize new objects by drawing upon a vast repository of prior knowledge while using the visual features of unfamiliar objects. Similarly, our approach combines the broad knowledge-base of LLMs with object-specific visual features from limited samples, resulting in a robust segmentation performance.
Our method addresses the limitations of prior work by providing richer contextual information and achieving superior performance on challenging benchmarks by a significant margin.
By integrating semantic and visual cues, it addresses FSS challenges and demonstrates LLMs' potential to improve segmentation and guide future research on complex tasks.

\section*{Acknowledgement}
This research was undertaken, in part, based on support from the Natural Sciences and Engineering Research Council of Canada Grants RGPIN-2021-03479 (NSERC DG) and ALLRP 571887 - 2021 (NSERC Alliance).

{\small
\bibliographystyle{ieeenat_fullname}
\bibliography{arxiv}}

\clearpage
\setcounter{page}{1}
\onecolumn
\begin{center}
    \Large\textbf{DSV-LFS: Unifying LLM-Driven Semantic Cues with Visual Features for Robust Few-Shot Segmentation}\\[1em]
    \large Supplementary Material
\end{center}

In the supplementary material, we provide qualitative examples with detailed class descriptions. Given the extensive number of images with full class descriptions, these are presented in a single-column format.




\section{Qualitative Results}

We present qualitative results from our proposed method to demonstrate its effectiveness in addressing key challenges in few-shot segmentation. The two primary challenges in this context are: (1) the misclassification of base class objects as novel classes, resulting in false positives, and (2) reliance on a limited set of support images, which often fails to capture the full range of target class variations.

The following examples illustrate challenging episodes that highlight these issues. In the examples, the input episode consists of annotated support and query images alongside the DSV-LFS output. While the support and query images are annotated to specify the object of interest, the annotation on the DSV-LFS output represents the predictions of the proposed method. Additionally, a detailed class description is provided as an input prompt for the multi-modal LLM to generate the semantic prompt.\textless Qimage\textgreater  in the class descriptions serves as a default token assigned to the query image within the input class description. This token is subsequently replaced by the output features of the query image.

\newpage

\label{sec:figure}
\begin{figure*}[ht]
    \centering
    \renewcommand{\arraystretch}{0.5} 
    \resizebox{\textwidth}{!}{ 
    \begin{tabular}{@{} c @{} c @{} c @{} c @{} c @{} c @{} c @{} c @{}}
        
        \parbox[c]{1.5cm}{\centering \scriptsize \textbf{Input Episode} \\ \textcolor{red}{Motorcycle}} &
        
        \begin{tabular}{c}
            \includegraphics[width=1.5cm]{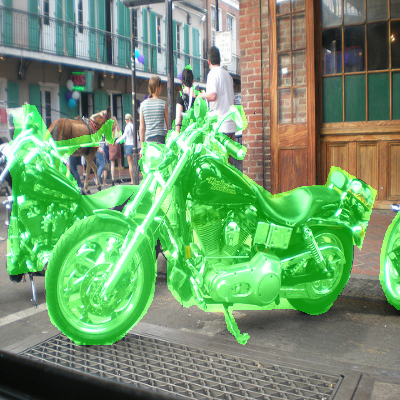} \\
            \parbox[t]{2.0cm}{\centering \scriptsize \tiny Support Image}
        \end{tabular} &
        
        \begin{tabular}{c}
            \includegraphics[width=1.5cm]{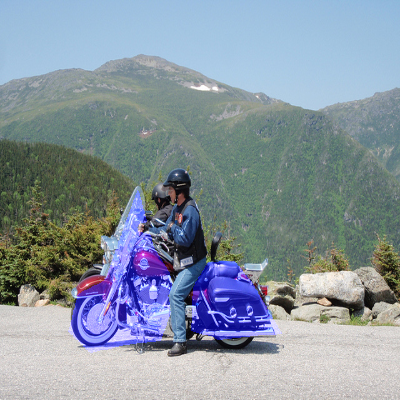} \\
            \parbox[t]{2.0cm}{\centering \scriptsize \tiny Query Image}
        \end{tabular} &

        \begin{tabular}{c}
            \includegraphics[width=1.5cm]{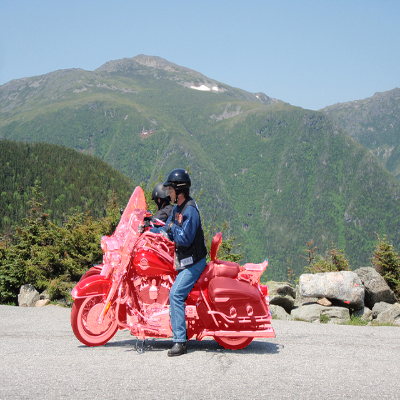} \\
            \parbox[t]{2.0cm}{\centering \scriptsize \tiny DSV-LFS}
        \end{tabular} \\

    \end{tabular}
    } 
    
    \label{fig:results}
\end{figure*}

\begin{tcolorbox}[
    colback=gray!20, 
    colframe=black, 
    sharp corners=south, 
    boxrule=0.75pt, 
    width=\textwidth, 
    arc=8pt, 
    auto outer arc, 
    boxsep=5pt, 
]

Complete class description for mutli-modal LLM for \textcolor{red}{motorcycle} object class:\\

\textcolor{blue}
{\textless Qimage\textgreater}. this one is a query image.
\textcolor{blue}{A motorcycle is distinctively characterized by its two-wheeled structure, which sets it apart from other vehicles. The wheels are large and typically exposed, with a prominent front wheel that often includes a visible brake disc and caliper, and a rear wheel that may have a broader tire. The frame is compact and streamlined, with a noticeable absence of a roof or any extensive enclosure. The handlebars, which are prominently situated above the front wheel, feature visible controls and mirrors extending outward. The seat is elongated and generally positioned for a straddling rider, often with a noticeable saddle shape. Beneath the seat, the engine is a dominant visual element, with its intricate metallic components like the exhaust pipes and cylinders often exposed. The fuel tank, typically located in front of the seat and above the engine, is a rounded or angular structure with a glossy finish. Additionally, motorcycles have a distinctive set of front and rear lights; the front light is usually a singular, circular or angular headlamp, while the rear includes a smaller brake light. The suspension system, including visible shock absorbers and forks, also adds to its unique visual identity. Overall, a motorcycle's open, mechanical design with exposed wheels, engine, and handlebars, along with its streamlined silhouette, distinctly sets it apart from bicycles, scooters, and other similar object categories.A motorcycle is distinguished from other similar object categories by several unique visual features. Primarily, it has two large wheels in tandem with a sleek, streamlined frame connecting them. The frame often exhibits a minimalistic, exposed design, typically featuring a prominent fuel tank situated between the handlebars and the riders seat. The handlebars, located at the front, are usually higher than the seat and are connected to a visible front fork that holds the front wheel. Unlike bicycles, motorcycles have a more robust structure with a bulkier engine block situated beneath the fuel tank, which is often exposed or partially covered. The exhaust system, consisting of one or more metal pipes, extends from the engine towards the rear. Motorcycles also feature foot pegs for the rider, often accompanied by additional pegs or a small seat for a passenger. The rear wheel is connected to the frame via a swingarm, which allows for suspension and typically houses a single large shock absorber. The design may include fairings, which are aerodynamic covers, though these are not as extensive as on scooters or other fully-enclosed vehicles. Additionally, motorcycles usually have larger, more prominent headlights and taillights compared to bicycles, often integrated into the design rather than being detachable. The tires on motorcycles are wider and more robust than those on bicycles, designed to handle higher speeds and more significant weight. These visual features collectively distinguish motorcycles from bicycles, mopeds, and scooters.} This paragraph outlines the visual features of motorcycle that distinguish it from other similar categories. Please use distinguishing visual features to segment motorcycle in the query image.

\end{tcolorbox}

\newpage

\begin{figure*}[ht]
    \centering
    \renewcommand{\arraystretch}{0.5} 
    \resizebox{\textwidth}{!}{ 
    \begin{tabular}{@{} c @{} c @{} c @{} c @{} c @{} c @{} c @{} c @{}}
        
        \parbox[c]{1.5cm}{\centering \scriptsize \textbf{Input Episode} \\ \textcolor{red}{Keyboard}} &
        
        \begin{tabular}{c}
            \includegraphics[width=1.5cm]{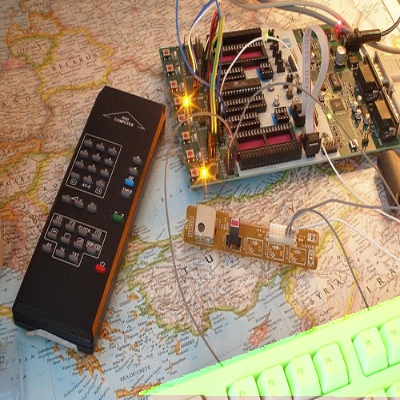} \\
            \parbox[t]{2.0cm}{\centering \scriptsize \tiny Support Image}
        \end{tabular} &
        
        \begin{tabular}{c}
            \includegraphics[width=1.5cm]{figures/WACVFIG/28/Qovrly.png} \\
            \parbox[t]{2.0cm}{\centering \scriptsize \tiny Query Image}
        \end{tabular} &

        \begin{tabular}{c}
            \includegraphics[width=1.5cm]{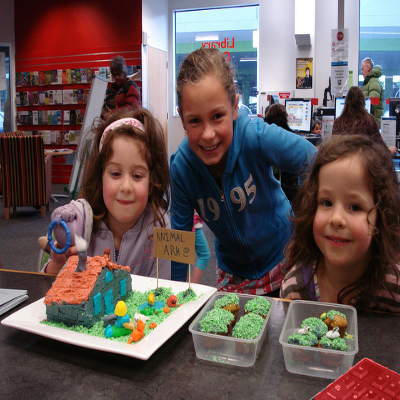} \\
            \parbox[t]{2.0cm}{\centering \scriptsize \tiny DSV-LFS}
        \end{tabular} \\

    \end{tabular}
    } 
    
    \label{fig:results}
\end{figure*}

\begin{tcolorbox}[
    colback=gray!20, 
    colframe=black, 
    sharp corners=south, 
    boxrule=0.75pt, 
    width=\textwidth, 
    arc=8pt, 
    auto outer arc, 
    boxsep=5pt, 
]

Complete class description for mutli-modal LLM for \textcolor{red}{keyboard} object class:\\

\textcolor{blue}
{\textless Qimage\textgreater}. this one is a query image.
\textcolor{blue}{A keyboard, in its distinctive visual form, is typified by its flat, elongated shape with an array of rectangular keys arranged in neat rows. Each key is typically square or slightly rectangular, often featuring rounded edges for ergonomic comfort during typing. The surface of the keys is uniformly smooth and matte or glossy, contrasting with the often darker or neutral-colored base. These keys are distinctly marked with alphanumeric characters, symbols, and functional indicators, often in contrasting colors such as white or light gray on dark backgrounds, aiding visibility and usability. Additionally, keyboards commonly include functional sections such as arrow keys, function keys (F1-F12), and a dedicated numerical keypad (on larger models), each section visually demarcated or slightly raised for tactile distinction. The overall profile of a keyboard is thin and flat, designed for ergonomic use on desks or tables, typically with a USB cable or wireless connectivity. These visual features collectively distinguish a keyboard from similar objects like calculators or remote controls, which lack the array of keys and alphanumeric layout essential for text input and control in computing environments.} This paragraph outlines the visual features of keyboard that distinguish it from other similar categories. Please use distinguishing visual features to segment keyboard in the query image.

\end{tcolorbox}

\begin{figure*}[ht]
    \centering
    \renewcommand{\arraystretch}{0.5} 
    \resizebox{\textwidth}{!}{ 
    \begin{tabular}{@{} c @{} c @{} c @{} c @{} c @{} c @{} c @{} c @{}}
        
        \parbox[c]{1.5cm}{\centering \scriptsize \textbf{Input Episode} \\ \textcolor{red}{Toilet}} &
        
        \begin{tabular}{c}
            \includegraphics[width=1.5cm]{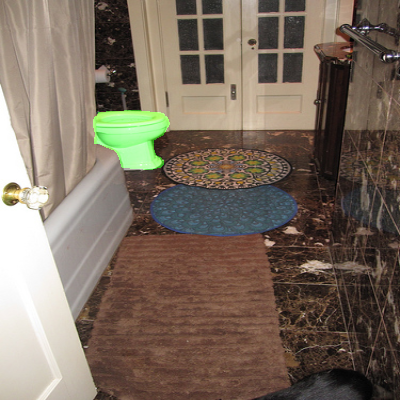} \\
            \parbox[t]{2.0cm}{\centering \scriptsize \tiny Support Image}
        \end{tabular} &
        
        \begin{tabular}{c}
            \includegraphics[width=1.5cm]{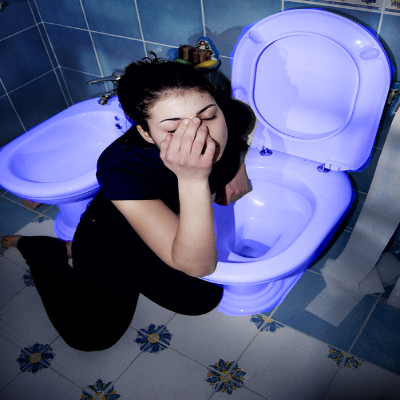} \\
            \parbox[t]{2.0cm}{\centering \scriptsize \tiny Query Image}
        \end{tabular} &

        \begin{tabular}{c}
            \includegraphics[width=1.5cm]{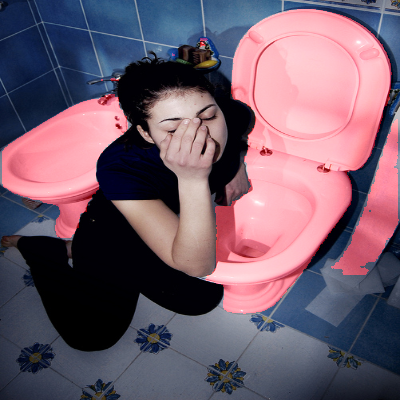} \\
            \parbox[t]{2.0cm}{\centering \scriptsize \tiny DSV-LFS}
        \end{tabular} \\

    \end{tabular}
    } 
    
    \label{fig:results}
\end{figure*}

\begin{tcolorbox}[
    colback=gray!20, 
    colframe=black, 
    sharp corners=south, 
    boxrule=0.75pt, 
    width=\textwidth, 
    arc=8pt, 
    auto outer arc, 
    boxsep=5pt, 
]

Complete class description for mutli-modal LLM for \textcolor{red}{toilet} object class:\\

\textcolor{blue}
{\textless Qimage\textgreater}. this one is a query image.
\textcolor{blue}{A toilet is a distinct bathroom fixture characterized by several unique visual features. It typically has a bowl-shaped seat made of porcelain or ceramic, with a rounded or oval opening that slopes inward. The bowl is often connected to a pedestal or base, which is relatively narrow compared to the bowl itself, giving it a recognizable silhouette. Attached to the back of the bowl is a water tank, which is usually rectangular and taller than it is wide, designed to hold flushing water. The toilet seat, often made of plastic, is hinged at the rear and can be lifted or lowered. This seat usually has a lid that matches in material and design. The bowl's interior is smooth and glazed, facilitating easy cleaning and often features a water-filled trap at the bottom, visible when the lid and seat are raised. The flush handle or button is typically located on the side or top of the water tank, which distinguishes it from other fixtures like bidets or urinals that lack such a tank. Overall, the combination of the bowl's shape, the attached water tank, the hinged seat and lid, and the flush mechanism make the toilet visually distinct from similar bathroom objects.}This paragraph outlines the visual features of toilet that distinguish it from other similar categories. Please use distinguishing visual features to segment toilet in the query image.

\end{tcolorbox}

\newpage
\begin{figure*}[ht]
    \centering
    \renewcommand{\arraystretch}{0.5} 
    \resizebox{\textwidth}{!}{ 
    \begin{tabular}{@{} c @{} c @{} c @{} c @{} c @{} c @{} c @{} c @{}}
        
        \parbox[c]{1.5cm}{\centering \scriptsize \textbf{Input Episode} \\ \textcolor{red}{Bird}} &
        
        \begin{tabular}{c}
            \includegraphics[width=1.5cm]{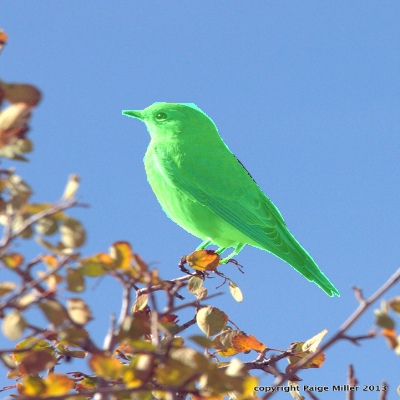} \\
            \parbox[t]{2.0cm}{\centering \scriptsize \tiny Support Image}
        \end{tabular} &
        
        \begin{tabular}{c}
            \includegraphics[width=1.5cm]{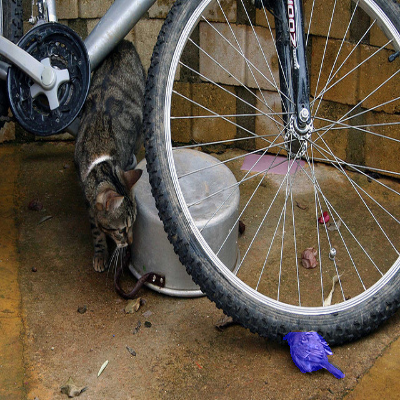} \\
            \parbox[t]{2.0cm}{\centering \scriptsize \tiny Query Image}
        \end{tabular} &

        \begin{tabular}{c}
            \includegraphics[width=1.5cm]{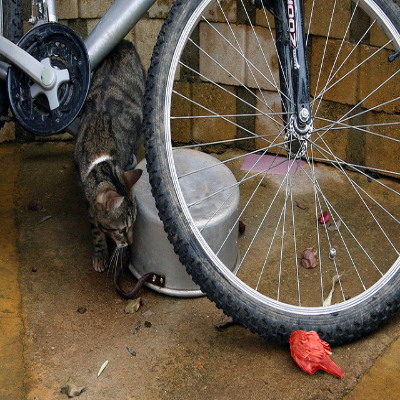} \\
            \parbox[t]{2.0cm}{\centering \scriptsize \tiny DSV-LFS}
        \end{tabular} \\

    \end{tabular}
    } 
    
    \label{fig:results}
\end{figure*}

\begin{tcolorbox}[
    colback=gray!20, 
    colframe=black, 
    sharp corners=south, 
    boxrule=0.75pt, 
    width=\textwidth, 
    arc=8pt, 
    auto outer arc, 
    boxsep=5pt, 
]

Complete class description for mutli-modal LLM for \textcolor{red}{bird} object class:\\

\textcolor{blue}
{\textless Qimage\textgreater}. this one is a query image.
\textcolor{blue}{Birds are characterized by their distinctive features, which set them apart from other similar object categories. Birds possess a unique feather covering, often brightly colored or patterned, providing insulation and aiding in flight. Their beaks vary in shape and size depending on their diet, from sharp, curved beaks in birds of prey to flat, broad ones in filter feeders. They have lightweight, streamlined bodies adapted for flight, with a high degree of symmetry and hollow bones. Their wings, a key identifier, exhibit a range of shapes and sizes, from long and narrow in soaring birds to short and rounded in those requiring rapid takeoff. The presence of a tail, often fan-shaped and used for steering during flight, further distinguishes them. Birds also have distinctive legs and feet, with variations such as webbed feet for swimming or strong talons for hunting. Their eyes are generally large and positioned on the sides of their heads, offering a wide field of vision. These combined features create a visual profile unique to birds, setting them apart from other animal categories.} This paragraph outlines the visual features of bird that distinguish it from other similar categories. Please use distinguishing visual features to segment bird in the query image.

\end{tcolorbox}

\begin{figure*}[ht]
    \centering
    \renewcommand{\arraystretch}{0.5} 
    \resizebox{\textwidth}{!}{ 
    \begin{tabular}{@{} c @{} c @{} c @{} c @{} c @{} c @{} c @{} c @{}}
        
        \parbox[c]{1.5cm}{\centering \scriptsize \textbf{Input Episode} \\ \textcolor{red}{Cow}} &
        
        \begin{tabular}{c}
            \includegraphics[width=1.5cm]{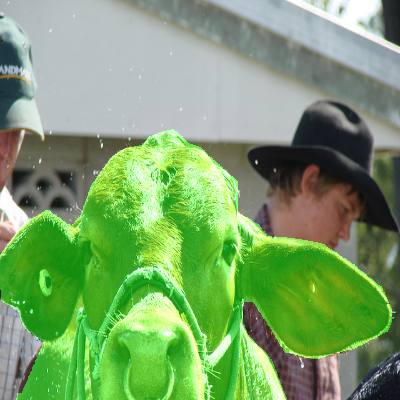} \\
            \parbox[t]{2.0cm}{\centering \scriptsize \tiny Support Image}
        \end{tabular} &
        
        \begin{tabular}{c}
            \includegraphics[width=1.5cm]{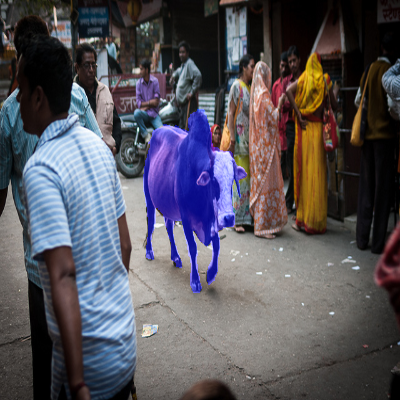} \\
            \parbox[t]{2.0cm}{\centering \scriptsize \tiny Query Image}
        \end{tabular} &

        \begin{tabular}{c}
            \includegraphics[width=1.5cm]{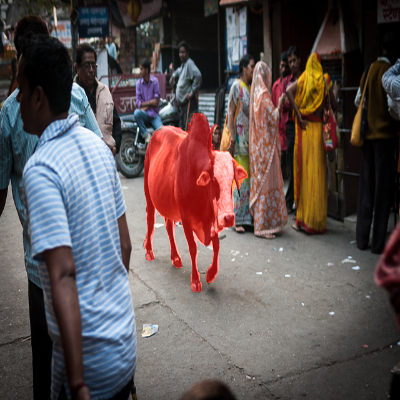} \\
            \parbox[t]{2.0cm}{\centering \scriptsize \tiny DSV-LFS}
        \end{tabular} \\

    \end{tabular}
    } 
    
    \label{fig:results}
\end{figure*}

\begin{tcolorbox}[
    colback=gray!20, 
    colframe=black, 
    sharp corners=south, 
    boxrule=0.75pt, 
    width=\textwidth, 
    arc=8pt, 
    auto outer arc, 
    boxsep=5pt, 
]

Complete class description for mutli-modal LLM for \textcolor{red}{cow} object class:\\

\textcolor{blue}
{\textless Qimage\textgreater}. this one is a query image.
\textcolor{blue}{Cows possess several distinguishing visual features that set them apart from similar object categories. They have a large, robust body with a pronounced rectangular shape, supported by four sturdy legs ending in cloven hooves. Their heads are relatively large, with broad, flat foreheads and distinctive long, broad snouts. A cow's eyes are large and round, usually positioned on the sides of their head, giving them a wide field of vision. They have large, prominent ears that can be either upright or slightly drooping. One of the most notable features is their pair of horns, which can vary in size and shape but are typically curved and symmetrical, though some cows may be naturally polled (hornless). Their tails are long and thin, ending in a tuft of hair, used to swat away insects. The skin of cows is covered in short hair, with color patterns that can vary significantly, including solid colors, spots, and patches in hues of black, white, brown, or a combination thereof. Unlike other similar animals, cows have a prominent udder with visible teats, particularly in dairy breeds, which is a key distinguishing feature. Additionally, cows have a distinctive gait and posture, often appearing more slow-moving and deliberate compared to other livestock.} This paragraph outlines the visual features of cow that distinguish it from other similar categories. Please use distinguishing visual features to segment cow in the query image.

\end{tcolorbox}

\newpage

\begin{figure*}[ht]
    \centering
    \renewcommand{\arraystretch}{0.5} 
    \resizebox{\textwidth}{!}{ 
    \begin{tabular}{@{} c @{} c @{} c @{} c @{} c @{} c @{} c @{} c @{}}
        
        \parbox[c]{1.5cm}{\centering \scriptsize \textbf{Input Episode} \\ \textcolor{red}{Hair dryer}} &
        
        \begin{tabular}{c}
            \includegraphics[width=1.5cm]{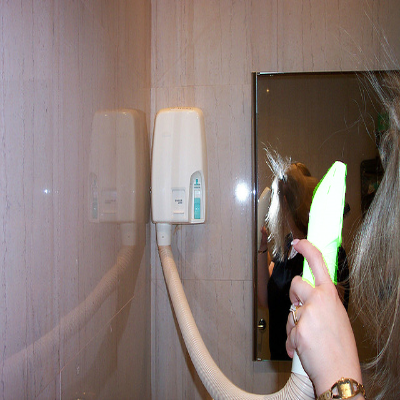} \\
            \parbox[t]{2.0cm}{\centering \scriptsize \tiny Support Image}
        \end{tabular} &
        
        \begin{tabular}{c}
            \includegraphics[width=1.5cm]{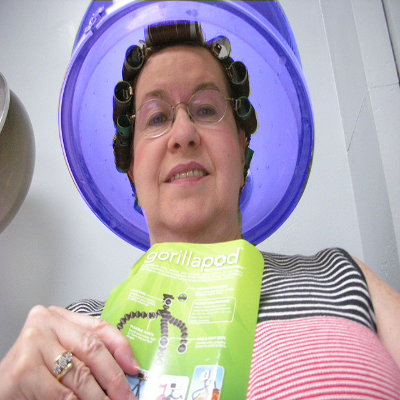} \\
            \parbox[t]{2.0cm}{\centering \scriptsize \tiny Query Image}
        \end{tabular} &

        \begin{tabular}{c}
            \includegraphics[width=1.5cm]{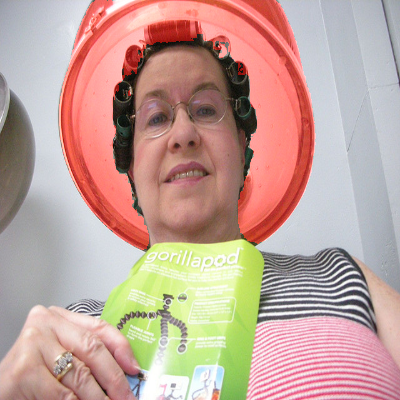} \\
            \parbox[t]{2.0cm}{\centering \scriptsize \tiny DSV-LFS}
        \end{tabular} \\

    \end{tabular}
    } 
    
    \label{fig:results}
\end{figure*}

\begin{tcolorbox}[
    colback=gray!20, 
    colframe=black, 
    sharp corners=south, 
    boxrule=0.75pt, 
    width=\textwidth, 
    arc=8pt, 
    auto outer arc, 
    boxsep=5pt, 
]

Complete class description for mutli-modal LLM for \textcolor{red}{hair dryer} object class:\\

\textcolor{blue}
{\textless Qimage\textgreater}. this one is a query image.
\textcolor{blue}{A hair dryer can be visually distinguished from similar objects primarily by its specific design features. Typically, a hair dryer consists of a cylindrical or slightly tapered body with a prominent handle and a nozzle at one end. The body often features a perforated grill or vents for airflow, which is essential for its function. The handle is ergonomically designed for grip and control, often contrasting in texture or color from the main body to enhance usability and visibility. On the body, there are frequently control buttons or switches for adjusting heat and airflow settings, which are clearly marked and distinct in appearance. The nozzle itself is narrow and elongated, sometimes with a distinct shape or curvature depending on the model, facilitating directional airflow during use. These visual characteristics collectively differentiate a hair dryer from other similar objects like handheld vacuum cleaners or electric razors, which have different body shapes, nozzle configurations, and control mechanisms tailored to their respective functions.} This paragraph outlines the visual features of hair dryer that distinguish it from other similar categories. Please use distinguishing visual features to segment hair dryer in the query image.

\end{tcolorbox}


\end{document}